\documentclass[ijoc,sglanonrev]{informs4_arXiv}
\usepackage{eqndefns-left} 
\RequirePackage{tgtermes}
\RequirePackage{newtxtext}
\RequirePackage{newtxmath}
\RequirePackage{bm}
\RequirePackage{endnotes}

\OneAndAHalfSpacedXII 

\usepackage{amsfonts}
\usepackage{indentfirst}
\usepackage{subfigure}
\usepackage{caption}
\usepackage{hyperref}
\usepackage{booktabs}
\usepackage{multirow}
\usepackage{threeparttable}
\usepackage{verbatim}
\usepackage{pifont}

\usepackage{algorithm}
\usepackage{algpseudocode}
\usepackage{tikz}

\usepackage{natbib}
 \bibpunct[, ]{(}{)}{,}{a}{}{,}%
 \def\bibfont{\small}%
\EquationsNumberedThrough    

\TheoremsNumberedThrough     
\ECRepeatTheorems  %

\begin{document}

\RUNAUTHOR{Wang, K., et al.}

\RUNTITLE{Tensor Completion Leveraging Graph Information}


\TITLE{Tensor Completion Leveraging Graph Information: A Dynamic Regularization Approach with Statistical Guarantees}

\ARTICLEAUTHORS{%
\AUTHOR{Kaidong Wang\textsuperscript{†}}
\AFF{School of Management, Xi'an Jiaotong University, Xi'an, China, \EMAIL{wangkd13@gmail.com}}

\AUTHOR{Qianxin Yi\textsuperscript{†}}
\AFF{School of Management, Zhengzhou University, Zhengzhou, China, \EMAIL{YiQianXin01@163.com}}

\AUTHOR{Yao Wang\textsuperscript{*}}
\AFF{School of Management, Xi'an Jiaotong University, Xi'an, China, \EMAIL{yao.s.wang@gmail.com}}

\AUTHOR{Xiuwu Liao}
\AFF{School of Management, Xi'an Jiaotong University, Xi'an, China, \EMAIL{liaoxiuwu@mail.xjtu.edu.cn}}

\AUTHOR{Shaojie Tang}
\AFF{School of Management, University at Buffalo, Buffalo, NY, USA, \EMAIL{shaojiet@buffalo.edu}}

\AUTHOR{Can Yang}
\AFF{Department of Mathematics, The Hong Kong University of Science and Technology, Hong Kong SAR, China, 
	\EMAIL{macyang@ust.hk}}

} 
\begingroup
\renewcommand{\thefootnote}{\fnsymbol{footnote}}
\footnotetext[2]{These authors contributed equally to this work.}
\footnotetext{\textsuperscript{*}Corresponding author.}
\endgroup
%
%
%

\ABSTRACT{
We consider the problem of tensor completion with graphs serving as side information to represent interrelationships among variables. Existing approaches suffer from several limitations: (1) they are often task-specific and lack generality or systematic formulation; (2) they typically treat graphs as static structures, ignoring their inherent dynamism in tensor-based settings; (3) they lack theoretical guarantees on statistical and computational complexity. To address these issues, we introduce a pioneering framework that systematically develops a novel model, theory, and algorithm for dynamic graph-regularized tensor completion. At the modeling level, we establish a rigorous mathematical representation of dynamic graphs and derive a new tensor-oriented graph smoothness regularization effectively capturing the similarity structure of the tensor. At the theory level, we establish the statistical consistency for our model under certain conditions, providing the first theoretical guarantees for tensor recovery in the presence of graph information. Moreover, we develop an efficient algorithm with guaranteed convergence.
A series of experiments on both synthetic and real-world data demonstrate that our method
achieves superior recovery  accuracy, especially under highly sparse observations and strong dynamics.
}%





\maketitle

\vspace{-5mm}
\section{Introduction}\label{introduction}
Over the past few years, low-rank tensor completion (TC) has attracted significant attention as an effective approach for recovering missing entries from partially observed high-dimensional data~\citep{cai2022nonconvex}. It has been successfully applied across diverse domains, including recommendation systems~\citep{farias2019learning}, biomedical data analysis~\citep{yi2022graph}, and intelligent transportation systems~\citep{sergin2025low}.
In the standard setting of third-order tensor data, the classical low-rank tensor completion problem seeks to recover a low-rank tensor $\mathcal{X} \in \mathbb{R}^{n_1 \times n_2 \times n_3}$ from a subset of observed entries $\mathcal{X}_{i_1 i_2 i_3}$, where $(i_1, i_2, i_3) \in \Omega \subset \{1,2,...,n_1\}\times \{1,2,...,n_2\}\times \{1,2,...,n_3\}$ and $|\Omega| \ll n_1 n_2 n_3$. A general formulation for TC is given by the rank minimization problem: 
$$ \min_{\check{\mathcal{X}}} \text{rank}(\check{\mathcal{X}})   ~~\text{s.t.} ~\check{\mathcal{X}}_{i_1i_2i_3} = \mathcal{X}_{i_1i_2i_3},  \forall (i_1, i_2, i_3)\in \Omega,$$
where the key challenge lies in defining an appropriate notion of tensor rank.
Over the past decades, various definitions of tensor rank have been proposed, including CP rank~\citep{du2019transit,bi2022improving}, Tucker rank~\citep{wang2023implicit}, tensor train (TT) rank~\citep{zhang2024tensor}, tensor ring (TR) rank~\citep{yu2025robust}. Each of these definitions offers distinct advantages and limitations. More recently, tensor singular value decomposition (t-SVD) and the induced tensor tubal rank~\citep{hou2021robust}  have gained increasing attention, owing to its advantages in the exploitation of  global structure information and spatial-shifting correlations inherent in tensor data.

In many real-world TC applications, relying solely on low-rankness often fails to yield satisfactory recovery, especially under highly limited observations. To address this, recent studies have explored the use of additional structural information, referred to as \textit{side information}, to improve recovery performance. Such side information generally falls into two categories: features, which represent attributes of individual entities (e.g., movie genres or directors)~\citep{farias2019learning, bertsimas2023interpretable}, and graphs, which describe relationships between entities (e.g., user social networks)~\citep{banerjee2016online}. Among the two, graphs often provide a more natural and flexible form of side information. This is primarily because: (1) feature-based completion methods typically require the data to lie within the feature space, which imposes a strong modeling assumption; and (2) features can often be naturally transformed into graphs. While feature representations may vary widely, it is generally reasonable to assume that entities with ``similar'' features exhibit stronger connectivity patterns in the associated graph. We therefore focus on TC with graph-based side information, where the graphs are either intrinsic to the data or constructed from feature-based similarity~\citep{rao2015collaborative}.

In recent years, several studies have explored matrix completion (MC) with graph information~\citep{kalofolias2014matrix, rao2015collaborative, dong2021riemannian}, demonstrating that incorporating graph structures can significantly improve recovery performance. However, extending such approaches to higher-order tensor completion introduces new challenges, primarily due to the dynamic nature of graphs in tensor data. In matrix settings, graphs are typically assumed to be static, remaining fixed once constructed. In contrast, tensor data often involves temporal  dimensions where relationships between entities evolve over time. For example, in a recommendation system, user social networks may change dynamically as new connections are formed and old ones fade.



The dynamic nature of graphs introduces two key challenges for tensor completion: mathematical modeling and theoretical analysis. To overcome these difficulties,, most existing methods adopt matrix-based strategies that rely on static graph assumptions. Specifically, they extend graph smoothness regularization from the matrix case by applying graph Laplacians directly to the tensor data. However, such approaches fail to respect the multi-dimensional structure of tensors and the temporal variability of graphs, often resulting in suboptimal recovery performance.
Moreover, most existing graph-regularized TC methods are designed for specific application scenarios. For instance, \citet{hu2020network} focuses on network latency estimation, \citet{yi2022graph} addresses microRNA-disease association prediction, and \citet{nie2023correlating} studies traffic data imputation. While effective in their respective domains, these methods lack a unified and generalizable framework, limiting their applicability across diverse tasks.
Furthermore, existing graph-regularized tensor completion methods generally lack theoretical guarantees, which undermines confidence in their performance and limits their reliability in real-world applications. This highlights the urgent need for a unified framework integrating dynamic graph information into tensor completion, supported by a well-defined mathematical model, provable theoretical guarantees, and an efficient optimization algorithm.

To address the above challenges, we identify four key research questions:
\begin{itemize}
\item How can dynamic graphs be represented in a rigorous mathematical form?
\item How can the similarity structures inherent in dynamic graphs be characterized holistically?
\item How can a unified model and efficient algorithm be developed to simultaneously exploit low-rank structure and dynamic graph similarity?
\item How can theoretical guarantees be established to ensure reliable recovery performance?
\end{itemize}
Building upon these four questions, we propose a unified framework for tensor completion with dynamic graph information, which systematically integrates a rigorous mathematical representation of dynamic graphs, a novel graph smoothness regularization, a comprehensive recovery model with an efficient algorithm, and a pioneering theoretical analysis establishing recovery guarantees.

In summary, the main contributions of this paper are as follows:
\begin{itemize}
	\item [1)]
	Taking into account the dynamic nature of graphs, we establish a rigorous mathematical representation for dynamic graphs. Based on this, we propose a novel tensor-oriented graph smoothness regularization that captures the global similarity structures embedded in the dynamic graph.
	\item [2)]
	We formulate a new model for TC with dynamic graph information, which integrates transformed t-SVD and the proposed graph smoothness regularization to jointly exploit the tensor’s low-rank and global similarity structures. An efficient ADMM-based algorithm with convergence guarantee is further developed to solve the resulting model.
	\item [3)]
	Employing a tensor atomic norm as a bridge, we establish the equivalence between the proposed graph smoothness regularization and a weighted tensor nuclear norm, and then derive statistical consistency guarantees for our model. To the best of our knowledge, this is the first theoretical guarantee for graph-regularized tensor recovery.
	\item [4)]
	We evaluate the proposed algorithm through extensive numerical experiments on both synthetic and real-world datasets. The results demonstrate the benefits of explicitly modeling dynamic graphs, and highlight the superior recovery performance of our method over several state-of-the-art baselines, especially under highly sparse observations and pronounced dynamics.
\end{itemize}

The remainder of this paper is organized as follows. In Section \ref{Related work} we provide a brief survey on the low-rank TC and MC/TC with graph information. The main notations and preliminaries are introduced in Section \ref{notations}. We present our main model in Section \ref{section model} and the optimization algorithm in Section \ref{section algorithm}. Consistency guarantees  of the proposed model is established in Section \ref{section theory}, and numerical experiments on synthetic and  real data are  conducted in Section \ref{section experiments}. Finally, in Section \ref{section conclusion}  we conclude
this paper and provide potential extensions of our method.

\subsection{Relevant Literature}\label{Related work}
\subsubsection{Low-rank Tensor Completion}
Over the past decade, a wide range of methods for low-rank tensor completion (TC) have been developed based on different tensor decomposition schemes and definitions of tensor rank. Classical approaches include CP and Tucker decomposition, where the former captures global multilinear structures through a superposition of rank-one tensors~\citep{cai2022nonconvex}, while the latter models mode-wise subspace interactions via a core tensor~\citep{wang2023implicit}.
More recently, tensor singular value decomposition (t-SVD) has gained significant attention due to its strength in capturing global structures and spatially shifting correlations commonly observed in real-world tensor data~\citep{hou2021robust}. Other strategies such as tensor train (TT) and tensor ring (TR)~\citep{zhang2024tensor, yu2025robust} have also been explored, but they often incur high computational overhead due to the need to reshape tensors into higher-order forms.

\subsubsection{Matrix/tensor Completion with Graph Information}
In recent years, several studies have incorporated graph information into matrix and tensor completion. For the matrix case, similarity graphs are typically modeled via graph Laplacian regularization and integrated into various matrix completion frameworks, including Gaussian processes~\citep{zhou2012kernelized}, nuclear norm minimization~\citep{kalofolias2014matrix}, and low-rank matrix decomposition~\citep{rao2015collaborative, dong2021riemannian}.
For tensors, most existing methods are task-specific, constructing static graphs tailored to particular applications. Hu et al.~\citep{hu2020network} incorporated node geometry into a graph Laplacian regularization term within a tensor nuclear norm minimization model for network latency estimation. Nie et al.~\citep{nie2023correlating} employed traffic networks as spatial regularization in TC for traffic speed estimation. Yi et al.~\citep{yi2022graph} used biological similarity network as  relational constraints to predict multiple types of microRNA-disease associations. In addition, Guan  et al.~\citep{guan2020alternating} proposed a general graph-regularized TC model  based on CP decomposition, incorporating graph Laplacian regularization to exploit static graph structures. These methods typically extend matrix-based regularization to tensor settings without accounting for the intrinsic dynamics of graph structures, and lack theoretical guarantees to support recovery performance.

Our work differs from prior works in two key aspects: (1) From a modeling perspective, our approach is the first to explicitly incorporate the intrinsic dynamics of graph structures, leading to significantly improved recovery performance. (2) From a theoretical perspective, we
provide a pioneering analysis of the statistical consistency for our model,
which is the first theoretical guarantees for graph regularized tensor recovery methods.

\section{Notations and Preliminaries}\label{notations}
\subsection{Notations}
We begin by summarizing the basic notations used throughout this paper. Scalars, vectors, matrices, and tensors are denoted by lowercase letters (e.g., $x$), bold lowercase letters (e.g., $\mathbf{x}$), uppercase letters (e.g., $X$), and calligraphic letters (e.g., $\mathcal{X}$), respectively. This work focuses on  three-order tensors, which are the most commonly encountered in practical applications. For a tensor $ \mathcal{X} \in \mathbb{R}^{n_1 \times n_2  \times n_3} $, its $ (i_1, i_2, i_3) $-th entry is denoted by $ \mathcal{X}_{i_1i_2i_3} $, its Frobenius norm is defined by $ \|\mathcal{X} \|_F = \sqrt{\sum_{i_1,i_2,i_3}\lvert \mathcal{X}_{i_1i_2i_3} \rvert ^2} $,  its infinity norm is defined by $ \|\mathcal{X}\|_\infty = \max_{i_1,i_2,i_3}\lvert \mathcal{X}_{i_1i_2i_3} \rvert $, and the inner product between two tensors $ \mathcal{X}, \mathcal{Y}  \in \mathbb{R}^{n_1 \times n_2  \times n_3} $ is defined by $ \left\langle  \mathcal{X}, \mathcal{Y} \right\rangle = \sum_{i_1,i_2,i_3} \mathcal{X}_{i_1i_2i_3}\mathcal{Y}_{i_1i_2i_3} $. Specifically, the $ i $-th
horizontal, lateral, and frontal slice of $ \mathcal{X} $ is denoted by $ \mathcal{X}_{i::}  $, $ \mathcal{X}_{:i:}  $ and $ \mathcal{X}_{::i}  $ (the frontal slice can also be denoted as $ \mathcal{X}^{(i)}  $ for brevity), and its $ (i,j) $-th tube fiber is denoted by $ \mathcal{X}_{ij:} $. 
\subsection{Tensor Singular Value Decomposition}
We briefly review the key definitions related to the transformed t-SVD.


For a tensor $ \mathcal{X} \in \mathbb{R}^{n_1 \times n_2 \times n_3} $, its mode-3 unfolding matrix is denoted by $ \mathcal{X}_{(3)}\in \mathbb{R}^{n_3\times n_1n_2} $. For any matrix $A \in \mathbb{R}^{m \times n_3}$, the tensor-matrix product  $ \mathcal{X} \times _3 A $ is defined by first  computing matrix-matrix product $ A\mathcal{X}_{(3)} $ and then reshaping the result to an $ n_1 \times n_2 \times m $ tensor. 
Let $ M \in \mathbb{R}^{n_3 \times n_3} $ be an invertible linear transform matrix satisfying $ MM^T=M^TM=CI $ for some constant $C $, where $ I $ is the
identity matrix. We define the mapping tensor of $ \mathcal{X}  $  in the transform domain as: $ \widehat{\mathcal{X}}: = \mathcal{X} \times _3 M $, and its inverse transform as $\widetilde{\mathcal{X}} := \mathcal{X} \times_3 M^{-1}$. Furthermore, we define $ \overline{\mathcal{X}} = \text{blockdiag}(\widehat{\mathcal{X}}) $ as the block diagonal matrix formed by stacking the frontal slices of $ \widehat{\mathcal{X}} $,
and its inverse operation is denoted by $ \text{fold}(\cdot) $: $ \text{fold}(\text{blockdiag}(\mathcal{X}))  = \mathcal{X} $. Then we can define the t-product of two tensors as follows:
\begin{definition}[tensor-tensor product (t-product)~\citep{kernfeld2015tensor}]
	The t-product of tensors $ \mathcal{A} \in \mathbb{R}^{n_1 \times n_2 \times n_3} $ and $ \mathcal{B} \in \mathbb{R}^{n_2 \times n_4 \times n_3} $ under invertible linear transform $ M\in \mathbb{R}^{n_3 \times n_3} $, denoted by $ \mathcal{A} * \mathcal{B} $, is a tensor $ \mathcal{C} \in \mathbb{R}^{n_1 \times n_4 \times n_3} $ given by: 
	$
		\mathcal{C} = \mathcal{A} * \mathcal{B} = \text{fold}(\overline{\mathcal{A}}\cdot\overline{\mathcal{B}})\times _3 M^{-1}.
	$
\end{definition}

We also use $ \star $ to denote the frontal slice-wise matrix multiplication between two tensors, i.e.,  $\mathcal{C} = \mathcal{A} \star \mathcal{B}$ is defined by $ \mathcal{C}^{(i)}= \mathcal{A}^{(i)}\mathcal{B}^{(i)},~ \text{for } i=1,2,...,n_3$.

We now introduce the definition of transformed t-SVD and several related concepts.
\begin{definition}[transformed t-SVD; tensor tubal rank \citep{kernfeld2015tensor}]
	For a tensor $ \mathcal{X} \in \mathbb{R}^{n_1 \times n_2 \times n_3} $, its transformed t-SVD is defined as:
	$
		\mathcal{X}  = \mathcal{U} * \mathcal{S}  * \mathcal{V}^T, 
	$
	where $ \mathcal{U} \in \mathbb{R}^{n_1 \times n_1 \times n_3} $, $ \mathcal{V} \in \mathbb{R}^{n_2 \times n_2 \times n_3} $ are unitary tensors, and $ \mathcal{S} \in \mathbb{R}^{n_1 \times n_2 \times n_3} $ 
	is a f-diagonal tensor. From this decomposition, the \emph{tubal rank} of $\mathcal{X}$ is defined as the number of nonzero tube fibers in $\mathcal{S}$.
\end{definition}
\begin{definition}[tensor spectral norm and nuclear norm~\citep{lu2019low}]
	The spectral norm and nuclear norm of a tensor $ \mathcal{X} \in \mathbb{R}^{n_1 \times n_2 \times n_3} $ under transform $ M $ is defined as $ \|  \mathcal{X} \| : = \|  \overline{\mathcal{X}} \| $ and $ \|  \mathcal{X} \|_* : = \frac{1}{C} \|  \overline{\mathcal{X}} \|_* $, respectively, where $\overline{\mathcal{X}} = \text{blockdiag}(\mathcal{X} \times_3 M)$, $ C $ is the constant for $ MM^T=CI $, and $\|\cdot\|$ and $\|\cdot\|_*$ denote the spectral norm and nuclear norm of a matrix, respectively.
\end{definition}

\begin{figure*}[t]
	\centering 
	\includegraphics[width=0.9\linewidth]{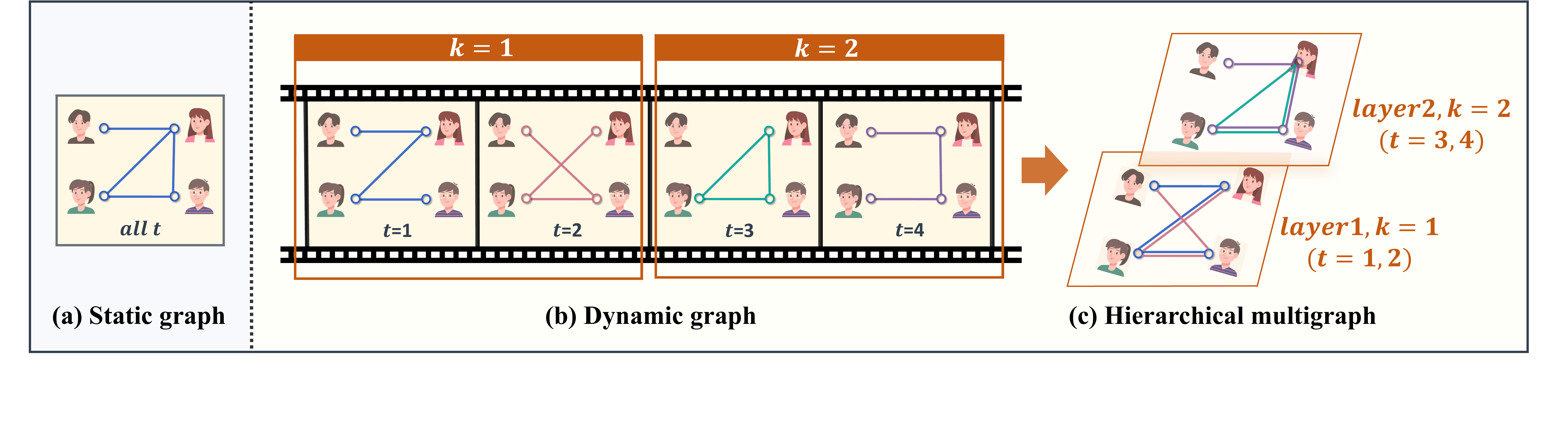}
	\vspace{-6mm}
	\caption{The illustration of (a) static graph, (b) dynamic graph with $ T=4 $ time periods and (c) corresponding hierarchical multigraph with $ K=2 $ and width of sliding time window $ \frac{T}{K} = 2 $, using a social network of users as the example.
	}\label{various graph}
	\vspace{-2mm}
\end{figure*}
\subsection{Problem Formulation}
In this subsection, we formulate a basic model for low-rank TC based on the transformed t-SVD framework. This choice is motivated by two key considerations: (1) t-SVD has shown superior capability in capturing global structural information in various tensor-based applications; (2) t-SVD preserves the tensor structure of the feature components, especially along the third mode (e.g., time), which facilitates the integration of dynamic graph information.

Suppose there exists a tensor $\mathcal{X} \in \mathbb{R}^{n_1 \times n_2 \times n_3}$, of which only a subset of entries $\mathcal{X}_{i_1 i_2 i_3}$, for $(i_1, i_2, i_3) \in \Omega$, are observed, where $|\Omega| \ll n_1 n_2 n_3$. The goal of TC is to estimate the unobserved entries $\mathcal{X}_{i_1 i_2 i_3}$ for $(i_1, i_2, i_3) \notin \Omega$. 
To exploit the underlying low-rank structure, we assume that $\mathcal{X}$ has low tubal rank $r^* \ll \min\{n_1, n_2\}$. Define the orthogonal projection operator $\mathcal{P}_\Omega(\cdot)$ that retains the observed entries in $\Omega$ and zeros out the rest, i.e., $ \mathcal{P}_\Omega(\mathcal{X})_{i_1i_2i_3}:= \mathcal{X}_{i_1i_2i_3} $ for $ (i_1,i_2,i_3) \in \Omega $ and $ 0 $ otherwise. 
Then, a basic low-rank tensor completion model can be formulated as:
\begin{equation}\label{tensor completion} 
	\begin{aligned}
		\check{\mathcal{W}}, \check{\mathcal{H}} = \arg\min_{\mathcal{W},\mathcal{H}} \frac{1}{2}\|\mathcal{P}_\Omega(\mathcal{X}-\mathcal{W}*\mathcal{H}^T)\|_F^2 + \frac{\lambda_1}{2}(\|\mathcal{W}\|_F^2 + \|\mathcal{H}\|_F^2),
	\end{aligned}
\end{equation}
where $\mathcal{W} \in \mathbb{R}^{n_1 \times r^* \times n_3}$ and $\mathcal{H} \in \mathbb{R}^{n_2 \times r^* \times n_3}$ are the factor tensors.
Instead of directly applying transformed t-SVD or minimizing the tubal rank, model~(\ref{tensor completion}) searches for a pair
of tensor factors with mode-2 dimension $ r^* $ to recover a tensor whose tubal rank is at most $r^*$.  
Since $r^* \ll \min\{n_1, n_2\}$, the sizes of $\mathcal{W}$ and $\mathcal{H}$ are significantly smaller than  the original tensor $\mathcal{X}$, resulting in substantial reductions in both computational cost and memory usage.

\section{Tensor Completion with Dynamic Graph Information}\label{section model}
In many scenarios, in addition to partially observed entries, additional information about relationships among entities, such as user social networks or product co-purchasing graphs, is available and can be leveraged to enhance prediction accuracy. Mathematically, such relationships can be naturally represented as a graph $(V, E)$, where $V$ and $E$ denote the sets of vertices and edges, respectively. For example, in a movie recommendation system, user ratings over $T$ time periods can be organized into a three-order tensor $\mathcal{X} \in \mathbb{R}^{m \times n \times T}$, where $m$ and $n$ denote the numbers of users and movies. Additional side information may include the users' social network or similarity among movies, both of which can be modeled as graphs: users (or movies) form the vertices, and their interactions or similarities define the edges.
Incorporating such graph-based information leads to a graph-regularized tensor completion formulation, which serves as the central focus of this work.

\subsection{Dynamic Graph Based Side Information }\label{side information}
As previously discussed, most existing studies adopt static graph representations, similar to those used in matrix completion, even when the underlying graphs in tensor-based applications may exhibit temporal dynamics. For instance, in collaborative filtering, a user's social connections may evolve over time as new relationships form and old ones fade.
To better capture such evolving structures, we introduce a dynamic graph representation, where the vertex set remains fixed while the edge set varies over time to reflect changing relationships among entities.
Figure~\ref{various graph} illustrates this concept with a user social network example: (a) presents a static graph, whereas (b) depicts a dynamic graph that reflects time-varying connections.

We now present a mathematical representation of dynamic graphs. A static graph with $m$ vertices can be described by a tuple $(V, E)$, where $V$ is the set of vertices and $E$ the set of edges. Its connectivity structure can be encoded by an adjacency matrix $A \in \mathbb{R}^{m \times m}$, defined as:
\begin{equation*} 
	A_{ij}:=\begin{cases} 1, \text{ if }(V_i, V_j) \in E, \text{ i.e., vertexes } V_i, V_j \text{ are adjacent}, \\ 0, \text{ if }(V_i, V_j) \notin E, \text{ i.e., vertexes } V_i, V_j \text{ are not adjacent},
	\end{cases}
\end{equation*}
To model evolving relationships, we represent a dynamic graph as a sequence of static graphs sharing the same vertex set, i.e., $(V, E_1), (V, E_2), \ldots, (V, E_T)$, with corresponding adjacency matrices $A_1, A_2, \ldots, A_T$ over $T$ time periods.
Based on this representation, we encode the time-varying adjacency relationships among the vertices in $V$ using a tensor $\mathcal{A} \in \mathbb{R}^{m \times m \times T}$ defined as:
\begin{equation*} 
	\mathcal{A}_{ijt}:=\begin{cases} 1, \text{ if }(V_i, V_j) \in E_t, \text{ i.e., vertexes } V_i, V_j \text{ are adjacent at time } t,
		\\ 0, \text{ if }(V_i, V_j) \notin E_t, \text{ i.e., vertexes } V_i, V_j \text{ are not adjacent at time } t. \end{cases}
\end{equation*}
We refer to $\mathcal{A}$ as the \textit{adjacency tensor} of the dynamic graph.

To facilitate our forthcoming discussion, we will employ a \textit{hierarchical multigraph} with the vertex set V as an equivalent representation of the dynamic graph. Differing from a simple graph, a multigraph is endowed with the ability to accommodate multiple edges, implying that two vertices can be connected by more than one edge. To create this hierarchical  multigraph from the dynamic graph, we first divide the dynamic graph into $ K $ disjoint continuous time intervals using a sliding time window with window width $ \frac{T}{K} $, each time interval corresponding to a dynamic subgraph. Denoting $ [k]: = [(k-1)\frac{T}{K}+1,(k-1)\frac{T}{K}+2,\cdots, k\frac{T}{K}]$ as the $ k $-th time interval, then the $ k $-th dynamic subgraph can  be represented with the adjacency subtensor $ \mathcal{A}_k = \mathcal{A}_{::[k]} $. For each dynamic subgraph, we construct a corresponding layer of the hierarchical  multigraph by retaining the vertices while sequentially introducing all the edges (repeated edges are considered as multiple edges) during each successive time period.  Consequently, we can obtain a hierarchical  multigraph with $ K $ layers, each layer aggregating the overall information of the corresponding dynamic subgraph. Apparently, when $ K=T $ the hierarchical  multigraph degenerates into the original dynamic graph, and when $ K=1 $ it degrades into an individual multigraph. Figure~\ref{various graph} (c) illustrates the corresponding hierarchical multigraph for the dynamic graph (b) on the left. Mathematically, a hierarchical multigraph can also be represented as the sets of its vertexes and all edges $ (V,E_1\cup E_2\cup \cdots) $ with adjacency tensor $ \breve{\mathcal{A}}\in \mathbb{R}^{m\times m\times K}  $ as
\begin{equation*} 
	\breve{\mathcal{A}}_{ijk}:=\sum_{t\in [k]} \mathcal{A}_{ijt} = \sum_{t\in [k]} \mathbb{I}((V_i, V_j) \in E_t), k=1,2,\cdots,K,
\end{equation*}
where $ \mathbb{I}(\cdot) $ is the indicator function. Meanwhile, for the convenience of subsequent derivations, based on adjacency tensor $ \breve{\mathcal{A}} $ and width of the sliding time window $ \frac{T}{K} $, we further define a corresponding elongated  adjacency tensor $ \vec{\mathcal{A}}\in \mathbb{R}^{m\times m\times T} $ as follows:
\begin{equation*}
	\vec{\mathcal{A}}_{ijt} = \mathcal{\breve{A}}_{ijk}, t\in [k],
\end{equation*}
which essentially duplicates each frontal slice of $ \breve{\mathcal{A}} $ into $ \frac{T}{K} $ identical slices.

\subsection{Tensor-oriented Graph Smoothness Regularization}
It is  known that in a matrix $ X = WH^T $ with graph $ (V,E) $ encoding the similarity between its rows  (or columns),  two rows connected by an edge in the graph
are ``close" to each other in the Euclidean distance. This property is usually called graph smoothness. Let $ L $ be the graph Laplacian matrix for $ (V,E) $, then the graph smoothness can be characterized by a regularizer 
\begin{equation}\label{matrix term}
	\text{Trace}(W^TLW) = \frac{1}{2}\sum_{i,j}E_{ij}(W_{i:}-W_{j:})^2,
\end{equation}
where $ W_{i:} $ and $ W_{j:} $ denote the $ i $-th and $ j $-th rows of $ W $, respectively. It can be seen that lessening (\ref{matrix term}) forces $ (W_{i:}-W_{j:})^2 $ to be small when $ E_{ij}=1 $, and thus the effect of graph smoothness can be achieved. However, in tensor cases the dynamic graph leads to the dynamic evolution of similarities, i.e., two individuals may be ``close'' to each other at some time periods while not at other periods. Thus, the key to tensor-oriented graph smoothness lies in how to model these dynamic similarities.

Taking the movie recommender system as an example, we consider the user–movie–time rating tensor $\mathcal{X} \in \mathbb{R}^{m \times n \times T}$, which is assumed to be low-rank and can be factorized as $\mathcal{X} = \mathcal{W} * \mathcal{H}^\top$, where $\mathcal{W} \in \mathbb{R}^{m \times r \times T}$ and $\mathcal{H} \in \mathbb{R}^{n \times r \times T}$ represent the user and movie feature tensors, respectively.
Beyond this, we obtain user (or movie) graph information in the form of a dynamic graph $\mathcal{G}$, which captures the evolving similarity among entities over time. A key characteristic of such a dynamic graph is its degree of dynamics: greater dynamics correspond to more rapid evolution of the similarity structure.
To adapt to this temporal variation, we define the notion of a \textit{similarity scale} ($ s $), which corresponds to the length of a time interval over which the graph structure remains relatively stable, and thus the similarity within the associated subtensor can be treated as a whole. Intuitively, a graph with stronger dynamics should be modeled using a smaller similarity scale, and vice versa.
As in Section \ref{side information}, using a sliding window of width $s = \frac{T}{K}$, the dynamic graph $\mathcal{G}$ can be represented as a hierarchical multigraph, with adjacency tensor $ \breve{\mathcal{A}}\in \mathbb{R}^{m\times m\times \frac{T}{s}} $ and elongated  adjacency tensor $ \vec{\mathcal{A}}\in \mathbb{R}^{m\times m\times T} $. 
Then we can extend the graph Laplacian matrix to the tensor setting by defining a f-diagonal tensor $ \vec{\mathcal{D}}  $ and  the corresponding graph Laplacian tensor as follows: 
\begin{equation*} 
	\vec{\mathcal{D}}_{ijt}:=\begin{cases} \sum_k  \vec{\mathcal{A}}_{ikt}, \text{ if } i=j, \\ 0, \text{~~~~~~~~~~~~ if } i \neq j, \end{cases}
	\mathcal{L}(\mathcal{G}, s) := \vec{\mathcal{D}} - \vec{\mathcal{A}}.
\end{equation*}
This construction preserves the temporal structure and generalizes the standard Laplacian to a dynamic graph context.

To capture the similarity structure of the user feature tensor $\mathcal{W}$ over the dynamic graph $\mathcal{G}$, we propose the following tensor-oriented dynamic graph smoothness regularization term:
\begin{equation}\label{regularization}
	\langle \widetilde{\mathcal{L}}(\mathcal{G}, s) , \mathcal{W} * \mathcal{W}^T  \rangle,
\end{equation}
where $ \widetilde{\mathcal{L}}(\mathcal{G}, s) $ denotes the mapping tensor of $ \mathcal{L}(\mathcal{G}, s) $ in the inverse linear transform domain.  To illustrate the intuition and justification of the regularization term~(\ref{regularization}), we reformulate it as an equivalent expression involving a product between the adjacency tensor of the hierarchical multigraph and pairwise distances within the tensor $\mathcal{W}$. As a result of rigorous derivation, $ \langle \widetilde{\mathcal{L}}(\mathcal{G}, s) , \mathcal{W} * \mathcal{W}^T  \rangle $  can be reformulated in the following equivalent form:
\begin{equation} \label{tensor regularization}
		\langle \widetilde{\mathcal{L}}(\mathcal{G}, s) , \mathcal{W} * \mathcal{W}^T  \rangle = \frac{1}{2}\sum_{i,j,k}\breve{\mathcal{A}}_{ijk} \| \mathcal{W}_{i:[k]}-\mathcal{W}_{j:[k]}\|_F^2,
\end{equation}
where $\breve{\mathcal{A}}$ denotes the adjacency tensor of the hierarchical multigraph, and $\mathcal{W}_{i:[k]}$ refers to a subslice of the horizontal slice $\mathcal{W}_{i::}$ with width $s$. From~(\ref{tensor regularization}), the proposed tensor-oriented graph smoothness regularization~(\ref{regularization}) can be interpreted as a weighted sum of pairwise distances between subslices $\mathcal{W}_{i:[k]}$ and $\mathcal{W}_{j:[k]}$, with weights given by the adjacency tensor $\breve{\mathcal{A}}^\mathcal{W}_{ijk}$ of the hierarchical multigraph. Minimizing~(\ref{tensor regularization}) encourages similar representations for users with frequent connections during a given time interval. The similarity scale $s$ controls the length of each interval, enabling adjustment of the temporal granularity in the dynamic graph. Hence,~(\ref{regularization}) provides a principled smoothness term that effectively integrates dynamic graph information into the tensor $\mathcal{W}$.


Based on the proposed tensor-oriented graph smoothness regularization, we develop a new model for tensor completion with dynamic graphs. Given an underlying tensor $\mathcal{X} \in \mathbb{R}^{n_1 \times n_2 \times n_3}$, partial observations $\mathcal{P}_\Omega(\mathcal{X})$, and dynamic graphs $\mathcal{G}^\mathcal{W}$ and $\mathcal{G}^\mathcal{H}$ for the two entity modes, we aim to jointly exploit the low-rank  and evolving graph structure in a unified framework.
By incorporating the regularization term~(\ref{regularization}) into the low-rank model~(\ref{tensor completion}), the proposed formulation can be written as:
\begin{equation}\label{model graph 2}
	\begin{aligned}\label{model graph 1}
		\check{\mathcal{W}}, \check{\mathcal{H}} 
		=\arg\min_{\mathcal{W},\mathcal{H}} \frac{1}{2}\|\mathcal{P}_\Omega(\mathcal{X}-\mathcal{W}*\mathcal{H}^T)\|_F^2+  \frac{1}{2}\big(\langle \mathcal{L}^\mathcal{W}, \mathcal{W} * \mathcal{W}^T  \rangle + \langle \mathcal{L}^\mathcal{H}, \mathcal{H} * \mathcal{H}^T  \rangle\big),
	\end{aligned}
\end{equation}
where $ \mathcal{L}^\mathcal{W} := \lambda_{\mathcal{G}}\widetilde{\mathcal{L}}(\mathcal{G}^\mathcal{W}, s) + \lambda_{1}\mathcal{I}^{\mathcal{W}} $ and $ \mathcal{L}^\mathcal{H} := \lambda_{\mathcal{G}}\widetilde{\mathcal{L}}(\mathcal{G}^\mathcal{H}, s) + \lambda_{1}\mathcal{I}^{\mathcal{H}} $, with $ \mathcal{I}^{\mathcal{W}}\in \mathbb{R}^{n_1\times n_1 \times n_3} $ and $ \mathcal{I}^{\mathcal{H}}\in \mathbb{R}^{n_2\times n_2 \times n_3} $ denoting the identity tensors. 
(\ref{model graph 2}) is our final model for TC with dynamic graphs. 

\section{Optimization Algorithm}\label{section algorithm}
In this section, we propose an efficient algorithm based on alternating direction method of multiplier (ADMM) and  conjugate gradient (CG) to solve the model (\ref{model graph 2}).  

\subsection{Optimization by ADMM}
We introduce auxiliary variables $\mathcal{A}$, $\mathcal{B}$, and $\mathcal{E}$, and rewrite model~(\ref{model graph 2}) in an equivalent form:
\begin{equation}\label{model2_2}
	\begin{aligned}
		&\min_{\mathcal{W},\mathcal{H}, \mathcal{A},\mathcal{B},\mathcal{E}} \frac{1}{2}\| \mathcal{E}-\mathcal{W}*\mathcal{H}^T\|_F^2
		+  \frac{1}{2}(\langle \mathcal{L}^\mathcal{W}, \mathcal{A} * \mathcal{A}^T  \rangle + \langle \mathcal{L}^\mathcal{H}, \mathcal{B} * \mathcal{B}^T  \rangle)\\
		&~~~~~~~~~\rm{s.t}.~\mathcal{A} = \mathcal{W}, \mathcal{B} = \mathcal{H},
		\mathcal{P}_\Omega(\mathcal{E}) = \mathcal{P}_\Omega(\mathcal{X})
		.
	\end{aligned}	
\end{equation}
The augmented Lagrangian function for (\ref{model2_2}) is:
\begin{equation}\label{model3_2}
	\begin{aligned}
		&L_A(\mathcal{W},\mathcal{H}, \mathcal{A},\mathcal{B},\mathcal{E})=\frac{1}{2}\|\mathcal{E}-\mathcal{W}*\mathcal{H}^T\|_F^2 +  \frac{1}{2}(\langle \mathcal{L}^\mathcal{W}, \mathcal{A} * \mathcal{A}^T  \rangle + \langle \mathcal{L}^\mathcal{H}, \mathcal{B} * \mathcal{B}^T  \rangle)\\
		&\quad \quad\quad+ \frac{\beta}{2} \| \mathcal{A} - \mathcal{W} - \frac{\bm{\lambda}^{\mathcal{W}}}{\beta} \|^2_F
		+ \frac{\beta}{2} \| \mathcal{B} - \mathcal{H}  - \frac{\bm{\lambda}^{\mathcal{H}}}{\beta} \|^2_F +   \frac{\beta}{2} \| \mathcal{P}_\Omega(\mathcal{E}) - \mathcal{P}_\Omega(\mathcal{X}) - \mathcal{P}_\Omega(\frac{\bm{\lambda}^{\mathcal{E}}}{\beta}) \|^2_F,
	\end{aligned}	
\end{equation}
where $\bm{\lambda}^{\mathcal{W}}$, $\bm{\lambda}^{\mathcal{H}}$  and $\bm{\lambda}^{\mathcal{E}}$ are the Lagrange multiplier tensors, and $ \beta $ is the positive penalty scalar. Then we can alternatively minimize $ L_A $ with respect to each variable by the following sub-problems.

\textbullet\ \textbf{$\mathcal{W},\mathcal{H}$ sub-problem: }The sub-problem to optimize $ L_A $ with respect to $ \mathcal{W} $ is:
$
	\min_{\mathcal{W}} \frac{\beta}{2} \| \mathcal{A} - \mathcal{W} - \frac{\bm{\lambda}^{\mathcal{W}}}{\beta} \|^2_F+ \frac{1}{2} \| \mathcal{E}-\mathcal{W}*_M\mathcal{H}^T\|^2_F,
$
which can be rewritten in the transform domain:
\begin{equation}\label{solve W_2_2}
	\min_{\widehat{\mathcal{W}}} \frac{\beta}{2C} \| \widehat{\mathcal{A}} - \widehat{\mathcal{W}} - \frac{\widehat{\bm{\lambda}^{\mathcal{W}}}}{\beta} \|^2_F+ \frac{1}{2C} \| \widehat{\mathcal{E}}-\widehat{\mathcal{W}}\star \widehat{\mathcal{H}^T} \|^2_F.
\end{equation} 
Considering the independence of those frontal slices in $ \widehat{\mathcal{W}}  $, $ \widehat{\mathcal{W}}^{(1)},...,\widehat{\mathcal{W}}^{(n_3)} $ can be solved separately and in parallel.  Denote a specific frontal slice $ \widehat{\mathcal{W}}^{(t)}$ as  $ \widehat{W}$, and  $ \widehat{\mathcal{A}}^{(t)}$ as  $ \widehat{A}$, $ \widehat{\mathcal{E}}^{(t)}$ as  $ \widehat{E}$, $ \widehat{\mathcal{H}^T}^{(t)}$ as  $ \widehat{H}^T$, $ \widehat{\mathcal{E}}^{(t)}$ as  $ \widehat{E}$,  $ \widehat{\bm{\lambda}^{\mathcal{W}}}^{(t)} $ as $ \widehat{\bm{\lambda}^{W}} $,  then solving $ \widehat{W}$ corresponds to the following optimization problem:
\begin{equation}\label{solve W_2_4}
	\min_{\widehat{W}}f(\widehat{W}) :=\frac{\beta}{2} \| \widehat{A} - \widehat{W}  - \frac{\widehat{\bm{\lambda}^{W}} }{\beta} \|^2_F+ \frac{1}{2} \| \widehat{E}-\widehat{W}\widehat{H}^T \|^2_F.
\end{equation} 
Optimization problem (\ref{solve W_2_4}) can be treated as solving the following linear system:
\begin{equation}\label{solve W_2_5}
	\widehat{W}(\widehat{H}^T\widehat{H}+\beta I)=\beta \widehat{A}-\widehat{\bm{\lambda}^{W}}+\widehat{E}\widehat{H},
\end{equation}
which can be solved by off-the-shelf CG techniques. After obtaining $ \widehat{\mathcal{W}} $, we can transform back to the original domain to get $ \mathcal{W} $. Obviously, $ \mathcal{H} $ can be solved in similar way as  $ \mathcal{W} $, except that  (\ref{solve W_2_5}) should be replaced by
\begin{equation}\label{solve H_2}
	\widehat{H}(\widehat{W}^T\widehat{W}+\beta I)=\beta \widehat{B}-\widehat{\bm{\lambda}^{H}}+\widehat{E}^T\widehat{W}
\end{equation}

\textbullet\ \textbf{$ \mathcal{A}, \mathcal{B} $ sub-problem: } Optimizing $ L_A $ with respect to $ \mathcal{A} $ can be formalized as:
$
	\min_{\mathcal{A}} \frac{\beta}{2} \| \mathcal{A} - \mathcal{W} - \frac{\bm{\lambda}^{\mathcal{W}}}{\beta} \|^2_F+  \frac{1}{2}(\langle \mathcal{L}^\mathcal{W}, \mathcal{A} * \mathcal{A}^T  \rangle).
$
We can also rewrite it in the transform domain, and solve $ \widehat{\mathcal{A}}^{(1)},...,\widehat{\mathcal{A}}^{(n_3)} $ separately and in parallel. Denote a specific frontal slice $ \widehat{\mathcal{A}}^{(t)}$ as  $ \widehat{A}$, and $\widehat{\mathcal{W}}^{(t)}$ as  $ \widehat{W} $, $ (\widehat{\mathcal{L}^\mathcal{W}})^{(t)} $ as $ \widehat{L^\mathcal{W}} $,  $ (\widehat{\bm{\lambda}^{\mathcal{W}}})^{(t)} $ as $ \widehat{\bm{\lambda}^{W}} $,  then solving $ \widehat{A} $ corresponds to the following optimization problem:
\begin{equation}\label{solve A_4}
	\min_{\widehat{A}} \frac{\beta}{2} \| \widehat{A}- \widehat{W} - \frac{\widehat{\bm{\lambda}^{W}} }{\beta} \|^2_F+  \frac{1}{2}\langle \widehat{L^\mathcal{W}} , \widehat{A}\widehat{A}^T  \rangle,
\end{equation}
which is equivalent to the linear system
\begin{equation}\label{solve A_5}
	(\widehat{L^\mathcal{W}}+\beta I)\widehat{A}=\beta \widehat{W} + \widehat{\bm{\lambda}^{W}}.
\end{equation}
We can solve $ \mathcal{B} $ in similar way apart from that (\ref{solve A_5}) should be replaced by 
\begin{equation}\label{solve B_5}
	(\widehat{L^\mathcal{H}}+\beta I)\widehat{B}=\beta \widehat{H} + \widehat{\bm{\lambda}^{H}}.
\end{equation} 
It is worth noting that  solving  $ \mathcal{A} $ and $ \mathcal{B} $ are independent and thus can be implemented in parallel.

\textbullet\ \textbf{$ \mathcal{E} $ sub-problem: } The sub-problem to optimize $ L_A $ with respect to $ \mathcal{E} $ has the following form:
\begin{equation}\label{solve E_2_1}
	\min_{\mathcal{E}}\frac{1}{2}\|\mathcal{E}-\mathcal{W}*_M\mathcal{H}^T\|_F^2 +   \frac{\beta}{2} \| \mathcal{P}_\Omega(\mathcal{E}) - \mathcal{P}_\Omega(\mathcal{X}) - \mathcal{P}_\Omega(\frac{\bm{\lambda}^{\mathcal{E}}}{\beta}) \|^2_F,
\end{equation}
Denoting $ \bar{\Omega} $  as the orthogonal complement of $ \Omega $, (\ref{solve E_2_1}) has the following explicit solution:
\begin{equation}\label{solve E_2_2}
	\begin{aligned}
		\mathcal{E}&=\frac{1}{1+\beta} \mathcal{P}_\Omega(\beta \mathcal{X}+\bm{\lambda}^{\mathcal{E}}+\mathcal{W}*_M\mathcal{H}^T) + \mathcal{P}_{\bar{\Omega}}(\mathcal{W}*_M\mathcal{H}^T).
	\end{aligned}
\end{equation}

\textbullet\ \textbf{Updating Multipliers: } Finally, the multipliers are updated by the following formulas:
\begin{equation}
	\label{update multipliers}
		\bm{\lambda}^{\mathcal{W}}\leftarrow \bm{\lambda}^{\mathcal{W}} - \gamma\beta (\mathcal{A} - \mathcal{W}), ~~
		\bm{\lambda}^{\mathcal{H}}\leftarrow \bm{\lambda}^{\mathcal{H}} - \gamma\beta \big(\mathcal{B} - \mathcal{H}),~~
		\bm{\lambda}^{\mathcal{E}}\leftarrow \bm{\lambda}^{\mathcal{E}} - \gamma\beta (\mathcal{P}_\Omega(\mathcal{E}) - \mathcal{P}_\Omega(\mathcal{X})),
\end{equation}
where $ \gamma $ is a parameter associated with convergence rate of the algorithm. 

Based on the above, the proposed algorithm for model~(\ref{model graph 2}) can now be summarized in Algorithm~\ref{algorithm1}.
\vspace{-8mm}
\begin{algorithm}[h]
	\caption{ Optimization Algorithm for the Model (\ref{model graph 2})}\label{algorithm1}
	\small
	\begin{algorithmic}[1]
		\State \textbf{Input:}  observed tensor $ \mathcal{P}_\Omega (\mathcal{X})\in 
		\mathbb{R}^{n_1\times n_2 \times n_3}$;  graphs $ \mathcal{G}^\mathcal{W}, \mathcal{G}^\mathcal{H} $; similarity scale $ s $;  tubal rank $ r $, parameters $ \lambda_{\mathcal{G}} $, $ \lambda_{1} $.
		\State \textbf{Output:} The reconstructed tensor $ \mathcal{X} = \mathcal{W}*\mathcal{H}^T $.
		\State \textbf{Initialization:}  Initial $ \mathcal{W}= \mbox{randn}(n_1,  r, n_3) $, $ \mathcal{H}= \mbox{randn}(n_2,  r, n_3) $, $ \mathcal{A} = \mathcal{W} $, $ \mathcal{B} = \mathcal{H} $, $ \mathcal{E}=\mathcal{P}_\Omega(\mathcal{X}) $.
		\While{not converge}
		\State
		Update $ \mathcal{W} $ via Eq.(\ref{solve W_2_5}).
		\State
		Update $ \mathcal{H} $ via Eq.(\ref{solve H_2});
		\State
		Update $ \mathcal{A} $, $ \mathcal{B} $ via Eq.(\ref{solve A_5}) and (\ref{solve B_5}), respectively;
		\State
		Update  $ \mathcal{E}$ via Eqs.(\ref{solve E_2_2}).
		\State
		Update multipliers via Eqs.(\ref{update multipliers}).
		\EndWhile
		\State  \textbf{Return}  $ \mathcal{X} = \mathcal{W}*\mathcal{H}^T $.
	\end{algorithmic}
\end{algorithm}
\vspace{-8mm}
\subsection{Computational Complexity}
We now analyze the computational complexity of Algorithm~\ref{algorithm1}. We use the Discrete Fourier Transform (DFT) in the t-SVD framework, resulting in a transform cost of $O(n_1 n_2 n_3 \log n_3)$ for a tensor of size $n_1 \times n_2 \times n_3$.  At each iteration, the main computational cost lies in solving subproblems (\ref{solve W_2_5}), (\ref{solve H_2}), (\ref{solve A_5}), (\ref{solve B_5}) and (\ref{solve E_2_2}). Solving~(\ref{solve W_2_5}) involves the tensor DFT and a conjugate gradient method for a linear system, with complexity $O(r^2\sqrt{K_1} + n_1 n_2 n_3 \log n_3)$, where $K_1$ is the condition number of the system. Similarly, the cost solving (\ref{solve H_2}), (\ref{solve A_5}), (\ref{solve B_5}) is $ O(r^2\sqrt{K_2} + n_1n_2n_3\log {n_3}) $, $ O(n_1^2\sqrt{K_3} + n_1^2n_3\log {n_3}) $, and $ O(n_2^2\sqrt{K_4} + n_2^2n_3\log {n_3}) $, respectively,  where $ K_2 $, $ K_3 $, $ K_4 $ are corresponding condition number. Solving~(\ref{solve E_2_2}) mainly involves tensor t-products and DFTs, with cost $O(r n_1 n_2 n_3 + r(n_1 + n_2) n_3 \log n_3)$. Summing all terms, the per-iteration complexity is  $ O(r^2\sqrt{K_1} + r^2\sqrt{K_2} +n_1^2\sqrt{K_3} + n_2^2\sqrt{K_4} + rn_1n_2n_3 +  ((n_1 + n_2)^2+r(n_1+n_2))n_3\log {n_3}) $. Assuming moderate condition numbers and $n_1 \approx n_2$, the complexity simplifies to $O(r n_1 n_2 n_3 + n_1 n_2 n_3 \log n_3)$.

\begin{figure}[t]
	\centering 
	\includegraphics[width=1\linewidth]{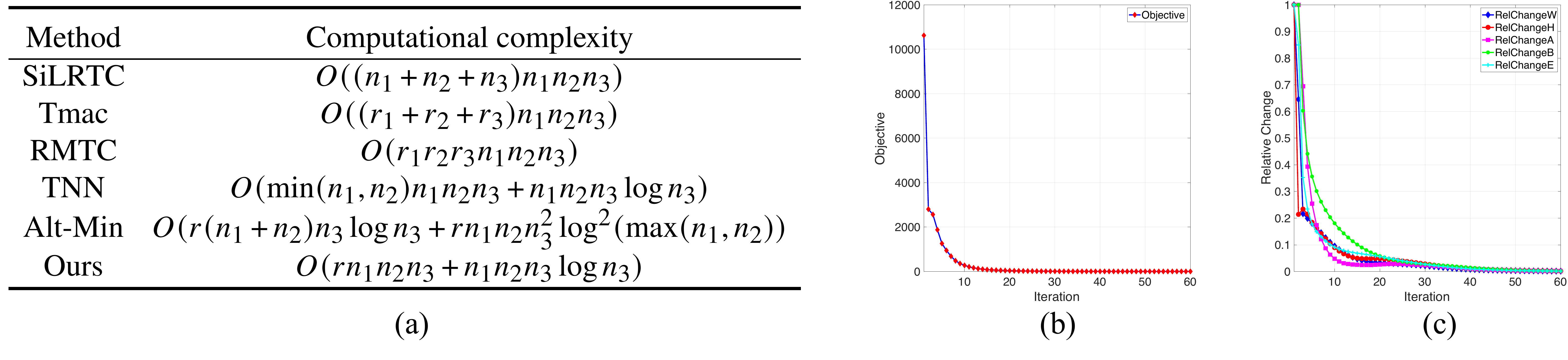}
	\vspace{-8mm}
	\caption{(a) Comparison of computational complexities of some representative TC methods. (b)(c) The objective and relative change curves with respect to the iterations of Algorithm~\ref{algorithm1}.
	}\label{figure_complexity_convergence}	
	\vspace{-4mm}
\end{figure}

In Figure~\ref{figure_complexity_convergence} (a), we compare the computational complexity of our method with several representative TC methods, including three  matricization-based methods SiLRTC~\citep{liu2012tensor}, Tmac~\citep{xu2013parallel}, RMTC~\citep{kasai2016low}, and two tensor-based methods TNN~\citep{zhang2014novel}, Alt-Min~\citep{liu2019low}, where $ r_1 $, $ r_2 $ and
$ r_3 $  denote the rank of the three unfolded
matrices used in matricization methods, respectively. As shown in Figure~\ref{figure_complexity_convergence} (a), our method achieves superior computational efficiency compared to these baselines.

\subsection{Algorithmic Convergence}
Denote the objective function of the optimization problem (\ref{model2_2}) as
\begin{equation}
	\begin{aligned}
		&F(\mathcal{W},\mathcal{H}, \mathcal{A},\mathcal{B},\mathcal{E})= \frac{1}{2}\| \mathcal{E}-\mathcal{W}*\mathcal{H}^T\|_F^2
		+  \frac{1}{2}(\langle \mathcal{L}^\mathcal{W}, \mathcal{A} * \mathcal{A}^T  \rangle + \langle \mathcal{L}^\mathcal{H}, \mathcal{B} * \mathcal{B}^T  \rangle),
	\end{aligned}	
\end{equation}
and denote the sequence generated by Algorithm~\ref{algorithm1} as $  (\mathcal{W}^k,\mathcal{H}^k,\mathcal{A}^k,\mathcal{B}^k,\mathcal{E}^k) $, then we can give the following convergence result of
Algorithm~\ref{algorithm1}.  The proofs of all the theorems presented later are provided in the supplementary material.

\begin{theorem}\label{theorem convengence}
			If $ \beta > r^2(n_1+n_2)n_3(\text{Trace}(\overline{\mathcal{L}^\mathcal{W}}) + \text{Trace}(\overline{\mathcal{L}^\mathcal{H}})) $, then the bounded sequence $ (\mathcal{W}^k,\mathcal{H}^k,\mathcal{A}^k,\mathcal{B}^k,\mathcal{E}^k) $ generated by
			Algorithm 1	satisfies the following properties:\\
			1.  Objective $F (\mathcal{W}^k,\mathcal{H}^k,\mathcal{A}^k,\mathcal{B}^k,\mathcal{E}^k) $ converges as $ k \rightarrow \infty $.\\
			2. The sequence $ (\mathcal{W}^k,\mathcal{H}^k,\mathcal{A}^k,\mathcal{B}^k,\mathcal{E}^k) $  has at least a limit point
			$ (\mathcal{W}^*,\mathcal{H}^*,\mathcal{A}^*,\mathcal{B}^*,\mathcal{E}^*) $, and any limit point $ (\mathcal{W}^*,\mathcal{H}^*,\mathcal{A}^*,\mathcal{B}^*,\mathcal{E}^*) $  is a feasible Nash point of the objective $ F $.\\
			3. For the sequence $ (\mathcal{W}^k,\mathcal{H}^k,\mathcal{A}^k,\mathcal{B}^k,\mathcal{E}^k) $, define $ u_k = \min_{0\leq l \leq k}(\|\mathcal{A}^{l+1}-\mathcal{A}^l\|_2^2+ \|\mathcal{B}^{l+1}-\mathcal{B}^l\|_2^2+\|\mathcal{W}^{l+1}-\mathcal{W}^l\|_2^2 + \|\mathcal{H}^{l+1}-\mathcal{H}^l\|_2^2 + \|\mathcal{P}_\Omega(\mathcal{E}^{l+1}-\mathcal{E}^l)\|_2^2) $, then the convergence
			rate of $ u_k $ is $ o(1/k) $.
\end{theorem}

Theorem~\ref{theorem convengence} establishes the convergence guarantee of the proposed algorithm, showing that it achieves a sublinear convergence rate of $o(1/k)$ despite the nonconvexity and complexity of the model. In addition, we provide empirical evidence to support its favorable convergence behavior. The convergence curves in Figure~\ref{figure_complexity_convergence} (b)(c) further validate the effectiveness of Algorithm~\ref{algorithm1}.
\section{Theoretical Analysis}\label{section theory}
In this section, we provide statistical guarantees for the proposed  model. To this end, we first demonstrate that the proposed graph smoothness regularization is actually equivalent to a weighted tensor nuclear norm, where the weights implicitly characterize the graph information.

\subsection{Equivalence to A Weighted Tensor Nuclear Norm}\label{equivalence}
In order to showcase the alignment between the proposed graph smoothness regularization and a weighted tensor nuclear norm, we utilize a tensor atomic norm as a conduit. To this end, we first generalize the definition of weighted atomic norm of matrices to the tensor case.  Let $ M \in \mathbb{R}^{T \times T} $ be a invertible linear transform matrix with $ MM^T = M^TM = CI $, and $ \mathcal{P}=  [\mathcal{P}_{:1:},\mathcal{P}_{:2:},...,\mathcal{P}_{:m:}] \in \mathbb{R}^{m \times m \times T} $ a basis of lateral slice space $ \mathbb{R}^{m \times 1 \times T} $ under transform matrix $ M $, i.e., each lateral slice $ \mathcal{W} \in \mathbb{R}^{m \times 1 \times T} $ can be expressed as a linear combination $ \mathcal{W} = \mathcal{P} * \mathcal{U} $ with representation $ \mathcal{U} \in \mathbb{R}^{m \times 1 \times T} $ and $ \|U\|_F = 1 $. Meanwhile, let $ \mathcal{S}_\mathcal{P} \in \mathbb{R}^{m \times m \times T} $ be a f-diagonal tensor with $ (\mathcal{S}_\mathcal{P})_{ii:} \geq 0, i=1,2,...,m $ encoding the weight of basis $ \mathcal{P} $ spanning the space. Obviously, tensor $ \mathcal{P} * \mathcal{S}_\mathcal{P} $ is also a basis. Similarly, let 
$ \mathcal{Q}=  [\mathcal{Q}_{:1:},\mathcal{Q}_{:2:},...,\mathcal{Q}_{:n:}] \in \mathbb{R}^{n \times n \times T} $ and $ \mathcal{S}_\mathcal{Q} \in \mathbb{R}^{n \times n \times T} $ be the basis and corresponding weight tensor for space $ \mathbb{R}^{n \times 1 \times T} $ under transform matrix $ M $. Denote $ \mathcal{A} = \mathcal{P} * \mathcal{S}_\mathcal{P} $ and $ \mathcal{B} = \mathcal{Q} * \mathcal{S}_\mathcal{Q} $, and we can then define a weighted tensor atomic set as follows:
\begin{equation}
	\begin{aligned}
		\mathscr{A} := \{\mathcal{D} = \mathcal{W} * \mathcal{H}^T: &\mathcal{W} = \mathcal{A} * \mathcal{U}, \mathcal{H} = \mathcal{B} * \mathcal{V}, \|\mathcal{U}\|_F = \|\mathcal{V}\|_F = 1\}.
	\end{aligned}
\end{equation}

\begin{definition}[weighted tensor atomic norm]
	Given a tensor $ \mathcal{X} \in \mathbb{R}^{m \times n \times T} $,  its weighted tensor atomic norm corresponding to the atomic set $ \mathscr{A} $ is: $ \|\mathcal{X}\|_{\mathscr{A}} := \text{inf} \sum_{\mathcal{D}_i \in \mathscr{A}} \lvert d_i \rvert, ~~ \text{s.t.} ~ \mathcal{X} =  \sum_{\mathcal{D}_i \in \mathscr{A}} d_i \mathcal{D}_i. $
\end{definition}

Based on those definitions, we can prove the following result:
\begin{theorem}\label{theorem 1}
	For any $ \mathcal{A} = \mathcal{P} * \mathcal{S}_\mathcal{P} $ and $ \mathcal{B} = \mathcal{Q} * \mathcal{S}_\mathcal{Q} $, and the corresponding weighted atomic set $ \mathscr{A} $, the following equality holds: 
			\begin{equation}
				\begin{aligned}
					\|\mathcal{X}\|_{\mathscr{A}} = \mathop{\rm inf}_{\mathcal{W},\mathcal{H}}  \frac{1}{2}\{\|\mathcal{A}^{-1} * \mathcal{W}\|_F^2 + \|\mathcal{B}^{-1} * \mathcal{H}\|_F^2 \}
					~~ \text{s.t.} ~ \mathcal{X} =  \mathcal{W} * \mathcal{H}^T. 
				\end{aligned}
	\end{equation} 
\end{theorem}

 Let $ \mathcal{L}^\mathcal{W} = \mathcal{U}_\mathcal{W} * \mathcal{S}_\mathcal{W} * \mathcal{U}_\mathcal{W}^T  $ and $ \mathcal{L}^\mathcal{H} = \mathcal{U}_\mathcal{H} * \mathcal{S}_\mathcal{H} * \mathcal{U}_\mathcal{H}^T  $ be the tensor eigen decomposition for $ \mathcal{L}^\mathcal{W} $ and $ \mathcal{L}^\mathcal{H} $. Define $ \mathcal{A} = \mathcal{U}_\mathcal{W} * \mathcal{S}_\mathcal{W}^{-1/2} $ and $ \mathcal{B} = \mathcal{U}_\mathcal{H} * \mathcal{S}_\mathcal{H}^{-1/2} $, then it can be verify that 
\begin{equation*}
	\begin{aligned}
		\langle \mathcal{L}^\mathcal{W}, \mathcal{W} * \mathcal{W}^T  \rangle =  \|\mathcal{A}^{-1} * \mathcal{W}\|_F^2, ~~ \langle \mathcal{L}^\mathcal{H}, \mathcal{H} * \mathcal{H}^T  \rangle = \|\mathcal{B}^{-1} * \mathcal{H}\|_F^2.
	\end{aligned}
\end{equation*}
By Theorem \ref{theorem 1}, the regularizers in model~(\ref{model graph 2}) are thus equivalent to a weighted tensor atomic norm.

Then we introduce two lemmas to show that the defined  weighted tensor atomic norm is actually a weighted tensor nuclear norm, and consequently
the equivalence between the regularizers in model (\ref{model graph 2})  and the weighted tensor nuclear norm can be directly obtained.  Given two weight tensors $  \mathcal{A} $ and  $ \mathcal{B} $, we can define a weighted tensor spectral norm of a tensor $  \mathcal{X} $ as $ \| \mathcal{X}\|_{ \mathcal{A},  \mathcal{B}} := \| \mathcal{A}^T * \mathcal{X} * \mathcal{B}\| $, based on which we introduce two lemmas.

\begin{lemma}\label{lemma 1}
	The dual norm of the weighted tensor spectral norm for a tensor $ \mathcal{X} $ is a weighted tensor nuclear norm: $ \| \mathcal{X}\|_{ \mathcal{A},  \mathcal{B}} ^* = \| \mathcal{A}^{-1} * \mathcal{X} * \mathcal{B}^{-T}\|_*.  $
\end{lemma}

\begin{lemma}\label{lemma 2}
	For $ \mathcal{A} = \mathcal{U}_\mathcal{W} * \mathcal{S}_\mathcal{W}^{-1/2} $ and $\mathcal{B} = \mathcal{U}_\mathcal{H} * \mathcal{S}_\mathcal{H}^{-1/2} $, and the corresponding weighted atomic set $ \mathscr{A} $, the dual norm of the weighted tensor atomic norm for $ \mathcal{X} $ is a weighted tensor spectral norm:
			\begin{equation*}
				\begin{aligned}
					\|\mathcal{X}\|_{\mathscr{A} ^*} &= \|\mathcal{X}\|_{ \mathcal{A},  \mathcal{B}} = \| \mathcal{A}^T * \mathcal{X} * \mathcal{B}\| = \|\mathcal{S}_\mathcal{W}^{-1/2} *  \mathcal{U}_\mathcal{W}^T * \mathcal{X} * \mathcal{U}_\mathcal{H} * \mathcal{S}_\mathcal{H}^{-1/2} \|.
				\end{aligned}
			\end{equation*}
\end{lemma}

By the two lemmas, the defined weighted tensor atomic norm is equivalent to a weighted tensor nuclear norm: $ \|\mathcal{X}\|_{\mathscr{A}} = \| \mathcal{A}^{-1} * \mathcal{X} * \mathcal{B}^{-T}\|_*. $
As a direct corollary, the regularization terms in model~(\ref{model graph 2}) are also equivalent to this weighted tensor nuclear norm, where the graph structures are implicitly encoded in the weight tensors $ \mathcal{A} = \mathcal{U}_\mathcal{W} * \mathcal{S}_\mathcal{W}^{-1/2} $ and $ \mathcal{B}= \mathcal{U}_\mathcal{H} * \mathcal{S}_\mathcal{H}^{-1/2} $.

\subsection{Main Theorem}
Based on the above conclusions, we then establish statistical consistency guarantees of the proposed dynamic graph regularized low-rank TC model.  We first make some assumptions and reformulate the problem. We assume that the true tensor $ \mathcal{X}^* \in \mathbb{R}^{m \times n \times T} $ is low rank with tubal rank $ r^*\ll \text{min}\{m,n\} $. 
We  denote the total number of entries in $ \mathcal{X}^* $ as $ D = m\times n \times T $,  and suppose that there are totally $ N \ll D$ noisy observations $ y_1, y_2, ..., y_N $ uniformly sampled from the following observation model:
$
	y_k = \langle \mathcal{X}^*, \mathfrak{X}_k \rangle + \sigma \xi _k, k=1, 2, ..., N, 
$
where $ \mathfrak{X}_k \in \mathbb{R}^{m \times n \times T} $ are i.i.d random design tensors;  random noise $ \xi _k \in \mathbb{R} $ are independent and centered sub-exponential variables with unit variance, and $ \sigma $  denotes the variance of noise. To specify the observation model in vector form, denote $ \bm{y} \in \mathbb{R}^N $ and $ \bm{\xi} \in \mathbb{R}^N $ as the observation and noise vectors, respectively, and meanwhile define a linear operator $ \mathscr{X}(\cdot): \mathbb{R}^{m \times n \times T} \rightarrow \mathbb{R}^N $ via
$
	\mathscr{X}(\mathcal{X})_k = \langle \mathcal{X}, \mathfrak{X}_k \rangle, k=1, 2, ..., N, \forall \mathcal{X} \in \mathbb{R}^{m \times n \times T}.
$
With those notations, we can then rewrite the observation model in vectorized form as: 
$
	\bm{y} = \mathscr{X}(\mathcal{X^*}) + \sigma \bm{\xi}. 
$

For the given $ \mathcal{L}^\mathcal{W} $ and $ \mathcal{L}^\mathcal{H} $, let $ \mathcal{A} $, $ \mathcal{B} $ be the corresponding weight tensors defined in Section~\ref{equivalence}, we then define a graph based complexity measure as
\begin{equation*}
	\alpha: =  \|\mathcal{A}^{-1} * \mathcal{X}^* * \mathcal{B}^{-T}\|_\infty.
\end{equation*}
Leveraging the established equivalence between the regularizers and a weighted tensor nuclear norm, model~(\ref{model graph 2}) can be reformulated as the following M-estimator:
\begin{equation}\label{estimator}
	\check{\mathcal{X}} = \arg\min_{\text{rank}(\mathcal{X})\leq r^*} \frac{1}{2N}\|\bm{y} - \mathscr{X}(\mathcal{X})\|_2^2 + \lambda \| \mathcal{A}^{-1} *_M \mathcal{X} *_M \mathcal{B}^{-T}\|_*,
\end{equation}
where $ \lambda $ is the trade-off parameter. We then analyze the statistical performance of estimator  (\ref{estimator}).

We first introduce a shorthand $ d = (m+n)T $ for brevity, and then a non-asymptotic upper bound on the Frobenius norm error is established in the following theorem.
\begin{theorem}\label{theorem 4}
	Suppose that we observe $ N $ entries from tensor $ \mathcal{X}^* \in \mathbb{R}^{m \times n \times T} $ under the above observation model, and $ \mathcal{X}^* $ has rank at most $ r^* $ with complexity measure $ \alpha $. Let  $ \check{\mathcal{X}} $ be the minimizer of the convex estimator (\ref{estimator}) with $ \lambda \geq c_0\sigma \sqrt{\frac{\log d}{Nn}} $, then we have 
			\begin{equation}
				\frac{\|\check{\mathcal{X}}  -\mathcal{X}^*  \|_F^2}{D} \leq  \alpha^2 c_1 \max \left \{  \max \{\sigma^2, 1\} \cdot \frac{r^*mT\log d}{N}, \frac{n\log d}{N} \right\}.
			\end{equation}
			with high probability, where $ c_0 $, $ c_1 $ are positive constants. 
\end{theorem}

Theorem~\ref{theorem 4} establishes the error bound of our model, where the influence of graph information is implicitly embedded in the measure $\alpha$. Theorem~\ref{theorem 4} guarantees that the per-entry estimation error of  our model satisfies 
$
	\frac{\|\check{\mathcal{X}}  -\mathcal{X}^*  \|_F^2}{mnT} \lesssim  \frac{\alpha^2r^*mT\log \left((m+n)T\right)}{N}
$
with high probability, where $ \lesssim $ means that the inequality holds up to a multiplicative constant. Equivalently,  it can be seen that with sample complexity $ O(\alpha^2r^*mT\log \left((m+n)T\right)) $, the per-entry estimation error will be small.

The following two remarks further demonstrate how incorporating dynamic graphs contributes to improving the recovery performance.

\begin{figure}[t]
	\centering 
	\includegraphics[width=1\linewidth]{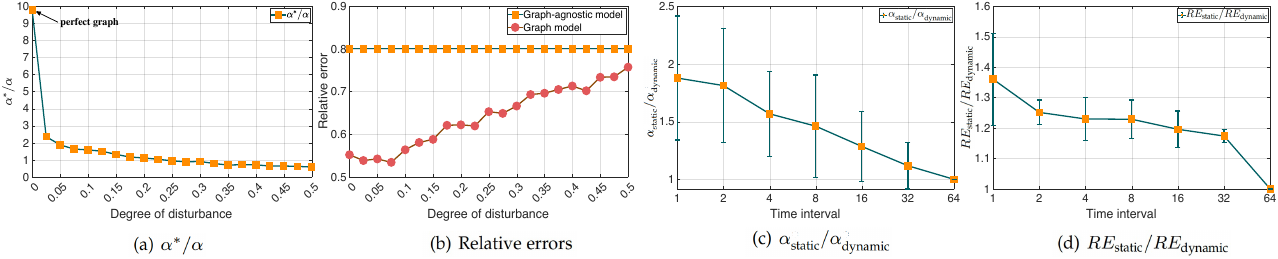}
	\vspace{-8mm}
	\caption{(a)(b) $  \alpha^*/\alpha  $ and relative recovery errors curves with respect to the degrees of perturbation in the graphs. (c)ratios of $  \alpha^* $ of the static version to original dynamic model under graphs with various time intervals;  (d) the corresponding ratios of relative errors.
	}\label{alpha_graphs}	
	\vspace{-4mm}
\end{figure}

\textbf{Remark 1.} By setting $ \mathcal{L}(\mathcal{G}^\mathcal{W}) = \mathcal{L}(\mathcal{G}^\mathcal{H}) = \mathbf{0} $, i.e., $ \mathcal{A}  =  \mathcal{B} = \mathcal{I}  $, model~(\ref{model graph 2}) reduces to a graph-agnostic version that ignores graph information. In this case, the corresponding error bound in Theorem~\ref{theorem 4} holds with $\alpha$ replaced by $\alpha^* := \|\mathcal{X}^*\|_\infty$.
We empirically demonstrate below by simulations that for informative graphs  $ \mathcal{G}^\mathcal{W} $ and $ \mathcal{G}^\mathcal{H} $,  $ \alpha $ is much smaller than $ \alpha^* $. We show the experimental settings in the Section 7 of supplementary material, and plot the corresponding $ \alpha^*/\alpha $ in Fig.~\ref{alpha_graphs} (a). From Fig.~\ref{alpha_graphs} (a) we can see that the perfect graph has the largest $ \alpha^*/\alpha $, and the ratio decreases as the disturbance in the graph increases. In Fig.~\ref{alpha_graphs} (b) we further compare the relative recovery errors of the graph-agnostic and original graph model, which shows that our original model can achieve smaller recovery errors than the graph-agnostic one, and the more informative graphs (i.e., the smaller $ \alpha $) lead to the smaller errors, which is consistent with the conclusion of Theorem \ref{theorem 4}.

\textbf{Remark 2.} By setting $s = T$ in model~(\ref{model graph 2}), the model reduces to a static variant that captures similarities across entire tensor slices and ignores temporal dynamics. 
To compare the $ \alpha $ of the static version and our original dynamic model, we denote their corresponding $ \alpha $ as $ \alpha_{\text{static}} $ and $ \alpha_{\text{dynamic}} $, respectively. We generate random tensors with size $ 50\times 50 \times 64 $ and corresponding dynamic graphs with various time intervals, the small time interval reflecting the more severe dynamics.  For each time interval, we compute the  $ \alpha $ and recovery relative error (RE) of the static model and dynamic model with $ s $ equaling to the time interval, respectively. We plot the curves  of $ \frac{\alpha_{\text{static}}}{\alpha_{\text{dynamic}}} $ and $ \frac{RE_{\text{static}}}{RE_{\text{dynamic}}} $ with respect to the time intervals in Fig~\ref{alpha_graphs} (c) and (d). From Fig~\ref{alpha_graphs} (c) we can see that $ \frac{\alpha_{\text{static}}}{\alpha_{\text{dynamic}}}\geq 1 $ and it increases as time interval decreases, which theoretically
verifies the superiority of  the dynamic model, especially in the  scenarios with severe graph dynamism. The $ \frac{RE_{\text{static}}}{RE_{\text{dynamic}}} $ curve in Fig~\ref{alpha_graphs} (d) also supports this point.


\section{Numerical Experiments}\label{section experiments}
\begin{figure}[t]
	\centering 
	\includegraphics[width=1\linewidth]{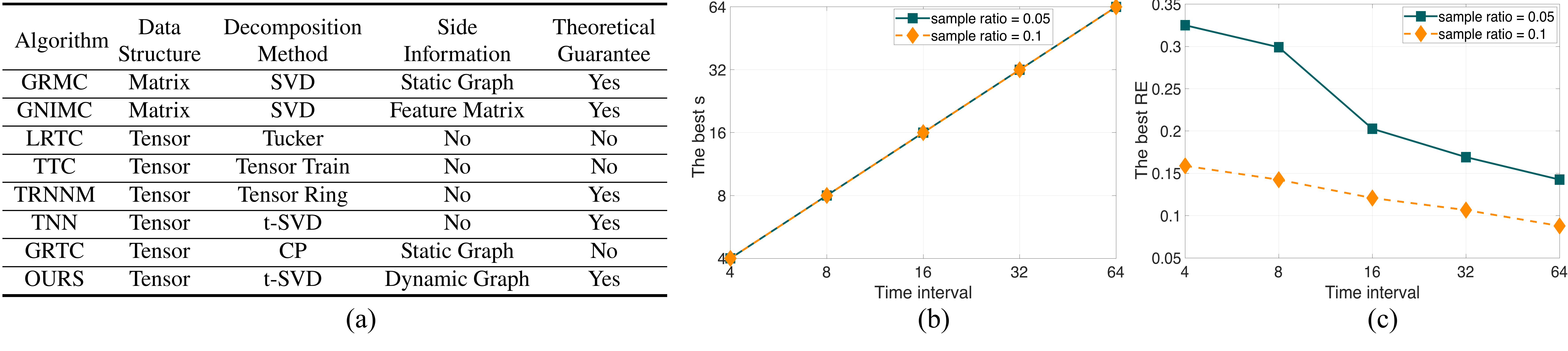}
	\vspace{-8mm}
	\caption{(a) Summary of some representative MC/TC models. (b)(c) Optimal similarity scales and corresponding relative errors for recovering tensors generated with dynamic graphs of varying time intervals.
	}\label{figure_similarityscale}	
	\vspace{-4mm}
\end{figure}

In this section, we evaluate the effectiveness of the proposed algorithm through a series of experiments performed on both synthetic and real-world datasets. The comparisons are made against several state-of-the-art matrix and tensor completion methods. Specifically, we include GRMC~\citep{dong2021riemannian} and GNIMC~\citep{zilber2022inductive} for matrix completion (MC), and LRTC~\citep{liu2012tensor}, TTC~\citep{chen2021auto}, TRNNM~\citep{huang2020provable}, TNN~\citep{lu2019low}, and GRTC~\citep{guan2020alternating} for tensor completion (TC). Details of the compared methods are summarized in Figure~\ref{figure_similarityscale} (a).   All experiments are implemented on a desktop computer with  Intel Core i9-9900k,
3.60 GHz, 64.0G RAM and  MATLAB R2022a. 

\subsection{Synthetic Data Experiments}\label{synthetic}
In this subsection we evaluate the performance of our algorithm on synthetic data. Specifically, we generate the ground-truth tensor $\mathcal{X} \in \mathbb{R}^{m \times n \times T}$ and corresponding dynamic graphs as follows.
To simulate dynamic graphs with varying levels of temporal variation, we assume that the graph structure changes at regular time intervals, with the interval length controlling the degree of dynamism. In each interval, a new graph is randomly generated using the “Community” model from the Graph Signal Processing Toolbox (GSPbox)~\citep{perraudin2014gspbox}. In this way, we generate two dynamic graphs $\mathcal{G}^\mathcal{W}$ and $\mathcal{G}^\mathcal{H}$ with corresponding Laplacian tensors $\mathcal{L}(\mathcal{G}^\mathcal{W})$ and $\mathcal{L}(\mathcal{G}^\mathcal{H})$.  Let $ \mathcal{L}(\mathcal{G}^\mathcal{W}) = \mathcal{U}_\mathcal{W} * \mathcal{S}_\mathcal{W} * \mathcal{U}_\mathcal{W}^T   $, $ \mathcal{L}(\mathcal{G}^\mathcal{H}) = \mathcal{U}_\mathcal{H} * \mathcal{S}_\mathcal{H} * \mathcal{U}_\mathcal{H}^T   $  be their tensor singular value decompositions. We then construct a low-rank tensor $\mathcal{X}$ with graph-induced smoothness as: $ \mathcal{Z} = \mathcal{P} * \mathcal{Q}^T, ~~ \mathcal{X} = \mathcal{A} \star \mathcal{Z} \star  \mathcal{B}^T $,
where $ \mathcal{P}\in \mathbb{R}^{m\times r\times T} $  and $ \mathcal{Q}\in \mathbb{R}^{n\times r\times T} $ are independently sampled from Gaussian distribution,  and tensors $  \mathcal{A} \in \mathbb{R}^{m\times m \times T} $ and $  \mathcal{B} \in \mathbb{R}^{n\times n \times T} $ are defined as
$
	\mathcal{A}  := \mathcal{U}_\mathcal{W} \star g(\mathcal{S}_\mathcal{W}),~~ \mathcal{B}  := \mathcal{U}_\mathcal{H} \star g(\mathcal{S}_\mathcal{H})
$
with graph spectral
filter $ g(\cdot) $. Here $ \mathcal{A} $ and $ \mathcal{B} $ transform the random matrix $ \mathcal{Z} $ into a graph smooth tensor $ \mathcal{X} $.

In the following experiments, we evaluate the recovery performance on a tubal rank 5 tensor $ \mathcal{X}\in \mathbb{R}^{50\times 50\times 64} $ with dynamic graphs $ \mathcal{G}^\mathcal{W}  $and $ \mathcal{G}^\mathcal{H}  $, where the degree of graph dynamism is controlled by varying the time interval length in $ \{4,8,16,32,64\} $. To this end, we randomly sample a proportion of entries from $\mathcal{X}$ as observations. The remaining entries are held out as the test set, and recovery performance is evaluated using the relative error defined as:
$
	\text{relative error (RE)} = \frac{\|\check{\mathcal{X}}_{\text{test}} - \mathcal{X}_{\text{test}} \|_F}{ \| \mathcal{X}_{\text{test}} \|_F},
$
where $ \check{\mathcal{X}} $ denotes the recovered tensor, and subscript $ \mathcal{X}_{\text{test}}  $ refers to the entries in the test set.

\subsubsection{Degree of Dynamism Versus Similarity Scale}\label{similarityscale}
In our model, the similarity scale ($s$) is introduced to adapt to different levels of graph dynamics. To examine the adaptability of $s$, we generate synthetic tensors along with dynamic graphs under varying time interval settings, where shorter intervals correspond to higher degrees of dynamism.
For each time interval setting, we evaluate the recovery performance of our model under different values of $s$. The value of $s$ that yields the lowest relative error is selected, and we plot the optimal $s$ and the corresponding relative errors as functions of the time interval in Fig.~\ref{figure_similarityscale} (b) and (c).  We observe the following: (1) the optimal similarity scale $s$ increases with the length of the time interval, indicating a clear positive correlation. This confirms that $s$ effectively adapts to varying levels of graph dynamism; (2) the corresponding relative error decreases as the time interval increases, which aligns with the intuition that more drastic temporal changes in the graph structure make tensor recovery more challenging.

\begin{figure}[t]
	\centering 
	\setlength{\abovecaptionskip}{0mm}
	\includegraphics[width=1\linewidth]{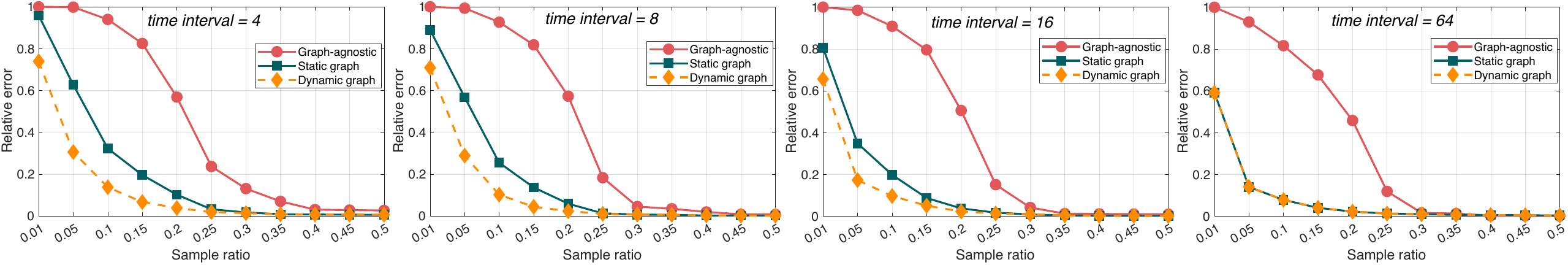}
	\caption{Performance of our model with (1) Graph-agnostic, (2) Static graph and (3) Dynamic graph model, for synthetic TC with graph information, where the sample ratios (the ratio of observed data)  are from $ 0.01 \sim 0.5 $, and the dynamic graphs are generated  with time intervals $ 4 $, $ 8 $, $ 16$, and $ 64 $.
	}\label{compare graph2}
	\vspace{-4mm}
\end{figure}
\subsubsection{Graph Comparison}\label{graphcomparison}
To assess the effectiveness of dynamic graph modeling and graph smoothness regularization in our framework, we compare the original model (referred to as \textit{Dynamic graph} model) against two degenerated baselines: (1) a general TC model without any graph information (\textit{Graph-agnostic} model), and (2) a TC model incorporating static graphs (\textit{Static graph} model).
Figure~\ref{compare graph2} shows the relative error curves of the three models across varying sample ratios and time intervals, averaged over five independent runs. From Fig.~\ref{compare graph2}  we can see that:\\
(1) In all four cases with different time intervals, both graph-based models consistently outperform the Graph-agnostic model, with the advantage more pronounced under lower sample ratios. This highlights the effectiveness of the proposed graph smoothness regularization in leveraging pairwise similarities encoded in the graphs to enhance recovery performance.  \\
(2) When the time interval is 64, the graphs $\mathcal{G}^\mathcal{W}$ and $\mathcal{G}^\mathcal{H}$ are  essentially  static, making the Dynamic and Static graph models equivalent. However, for time intervals of 4, 8, or 16, the Dynamic graph model consistently achieves lower relative errors, with the performance gap increasing as the time interval decreases. This demonstrates the benefit of explicitly modeling graph dynamics, which the Static graph model fails to capture.	

\subsubsection{Comparison with State-of-the-arts}
\begin{figure}[t]
	\centering 
	\setlength{\abovecaptionskip}{0mm}
	\includegraphics[width=1\linewidth]{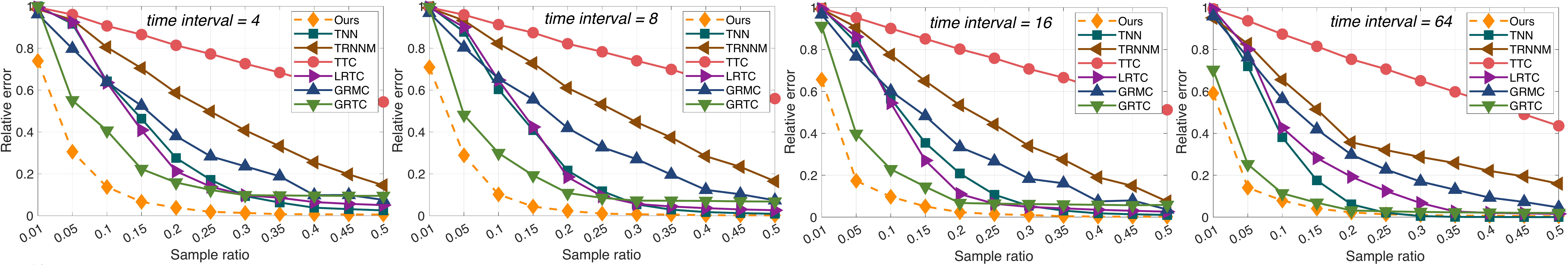}
	\caption{Performance of the compared methods for synthetic TC with graph information, where the sample ratios are from $ 0.01 \sim 0.5 $, and the dynamic graphs are generated  with time intervals $ 4 $ , $ 8 $ , $ 16$, and $ 64 $.
	}\label{compare synthetic}
	\vspace{-4mm}
\end{figure}

We evaluate our model against several representative MC/TC methods, including GRMC, LRTC, TTC, TRNNM, TNN, and GRTC. LRTC, TTC, TRNNM, and TNN recover tensors without using any graph information. GRMC is a graph-based MC method, adapted here by applying it to each frontal slice separately. GRTC uses static graphs, and we adopt the graphs from the first time interval in $\mathcal{G}^\mathcal{W}$ and $\mathcal{G}^\mathcal{H}$ for it. GNIMC is excluded due to the absence of feature matrices in the synthetic data.
We fix our parameters at $\lambda_{\mathcal{G}} = \lambda_1 = 0.001$, and tune baselines via recommended settings or cross-validation. The average relative errors under different sample ratios and time intervals are shown in Figure~\ref{compare synthetic}.
From Figure~\ref{compare synthetic}, it can be seen that:\\
(1) Among the four general TC methods, TRNNM and TTC consistently perform poorly across all time intervals. While TNN and LRTC achieve reasonable accuracy at high sample ratios, their performance degrades rapidly as the ratio decreases. This is because these methods rely solely on low-rank structures and fail to leverage the underlying similarity patterns.\\
(2) As a MC method, GRMC is inherently limited when applied to tensor recovery. Nevertheless, it performs comparably to some tensor-based methods, owing to its effective use of graph information.\\
(3) By leveraging static graphs, GRTC outperforms general TC methods, demonstrating the utility of graph-based similarity in enhancing tensor recovery. However, its performance degrades notably as the time interval shortens, highlighting its limitations in capturing dynamic graph structures.\\
(4) When the time interval is 64, the graph is effectively static, and our method achieves comparable performance to GRTC while outperforming all other baselines. In more dynamic scenarios (time intervals = 4, 8, 16), our method significantly outperforms all compared methods and exhibits strong adaptability to varying degrees of graph dynamics, primarily due to its effective integration of dynamic graph information.
\subsection{Real-World Experiments: Collaborative Filtering}
In this subsection, we evaluate the proposed method on a real-world application: collaborative filtering (CF), a widely adopted approach for modeling user behavior and building recommendation systems. CF typically involves constructing a $ user \times item $ rating matrix or a $ user \times item \times context $ tensor, and predicting missing entries from partially observed data using MC or TC techniques.

\subsubsection{Dataset Introduction}
We conduct experiments on the widely used benchmark dataset MovieLens-1M~\citep{harper2015movielens}, which contains $ 1,000,000 $ movie ratings from 6040 users on 3900 movies, along with user and movie feature information. Each rating is associated with a timestamp, and each user/movie is represented by a 22/18 dimensional feature vector, respectively. By splitting the timeline into 6 periods, we construct a sparse tensor of size $6040 \times 3900 \times 6$, with feature matrices available along the first two modes. To reduce computational cost, we randomly select $100$, $500$, and $1000$ users and movies to form three target tensors for evaluation.

\subsubsection{Graph Construction and Comparison}\label{Graph Construction}
Since the dataset does not provide graph information, we construct artificial user-wise and movie-wise similarity graphs. Specifically, we augment the raw features with artificial ones, and compute pairwise distances to build both static and dynamic graphs. Figure~\ref{graph_ss} (b) compares the performance across different graph models, illustrating the effectiveness of dynamic graph modeling.


\subsubsection{Comparison with State-of-the-arts}\label{movielens compare}
\begin{figure}[t]
	\centering 
	\includegraphics[width=1\linewidth]{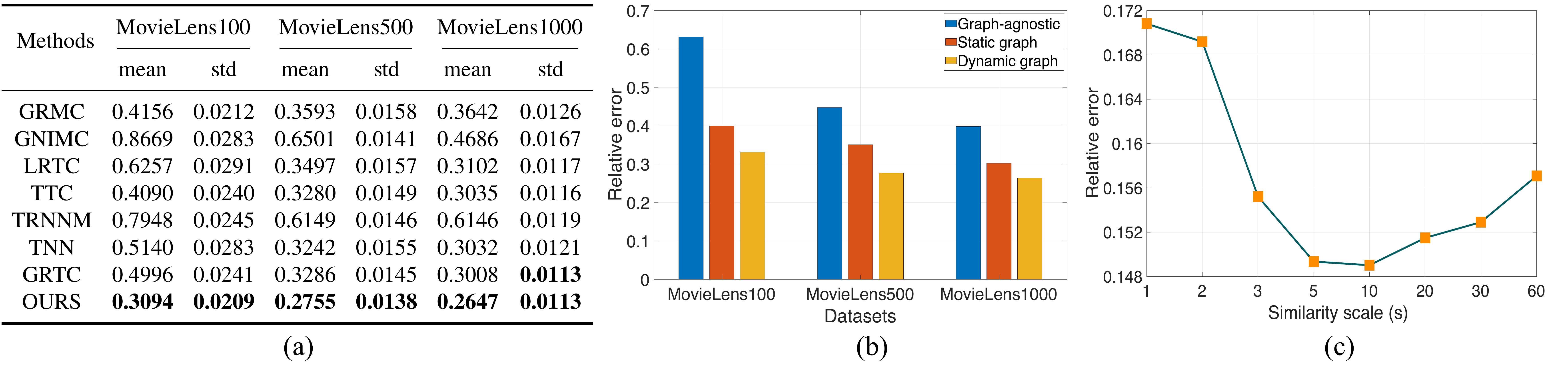}
	\vspace{-8mm}
	\caption{(a) Recovery performances of the tested methods on the MovieLens dataset; (b) Relative errors of various graph models in MovieLens dataset; (c) Relative errors with respect to similarity scales (s) in GuangZhou dataset.
	}\label{graph_ss}	
	\vspace{-4mm}
\end{figure}

We evaluate the recovery performance of our model against the aforementioned MC/TC methods on all three datasets. For each dataset, 80\% of the known ratings are randomly selected as training data, with the remaining 20\% used for testing. In our model, the parameters are fixed as $ \lambda_{\mathcal{G}} = \lambda_1 = 0.001 $, and the tubal rank is chosen via five-fold cross validation. For the baselines, parameters are either set to recommended values or selected similarly. Each experiment is repeated five times, and the average relative errors with standard deviations are reported in Figure~\ref{graph_ss} (a). The results show that our method consistently achieves the lowest errors and variances across all datasets, highlighting its robustness and superior performance.
\subsection{Real-World Experiments: Spatiotemporal Traffic Data Imputation}
In this subsection, we validate our proposed method on the task of spatiotemporal traffic data imputation, where traffic states (e.g., speed or link volume) are typically organized into tensor structures, and missing entries are estimated using tensor completion techniques.

\subsubsection{Dataset Introduction}\label{Dataset Introduction}
We evaluate the recovery performance of the proposed method on two spatiotemporal traffic datasets: GuangZhou\footnotemark[1]\footnotetext[1]{\href{https://zenodo.org/record/1205229}{https://zenodo.org/record/1205229}} and Portland\footnotemark[2]\footnotetext[2]{\href{https://portal.its.pdx.edu/home}{https://portal.its.pdx.edu/home}}.
The GuangZhou dataset contains traffic speed observations collected from 214 road segments in Guangzhou, China, over two months at 10-minute intervals, which can be originzed as a third-order tensor of size $214 \times 144 \times 60$ ($ road\ segment \times time\ interval \times day$). The Portland dataset collects the link volume from highways in Portland, USA containing 1156 loop detectors within one month at 15-minute interval, from which we construct a tensor ($ loop\ detectors \times time\ interval \times day  $)  with size $ 80\times 96 \times 15 $ to decrease computational overlead. As the two datasets lack explicit graph information, we construct artificial similarity graphs by first extracting features from the first two modes via t-SVD decomposition, followed by computing pairwise Euclidean distances to generate the graph structures.

\begin{figure}[t]
	\centering 
	\includegraphics[width=1\linewidth]{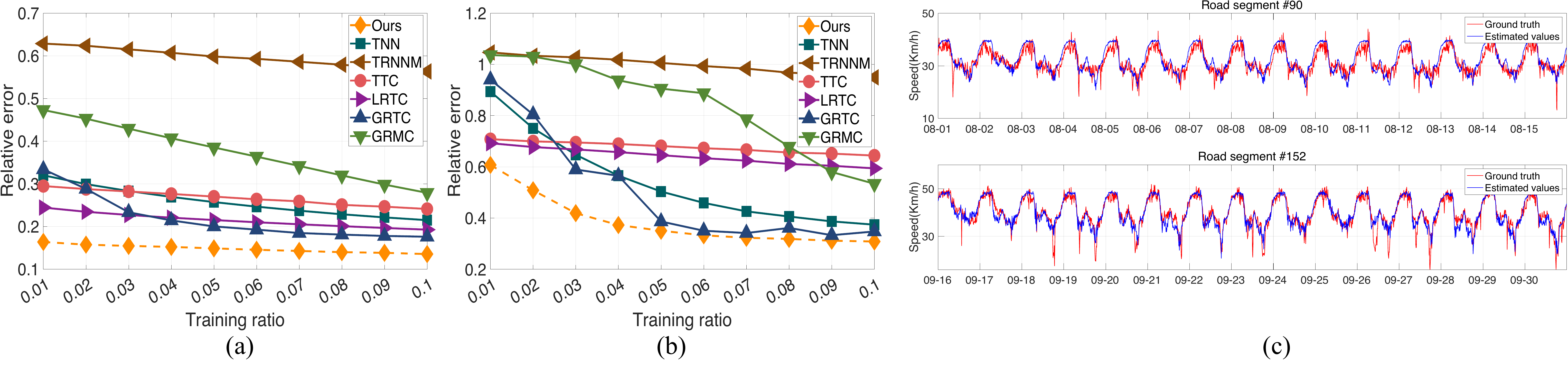}
	\vspace{-8mm}
	\caption{(a)(b) Recovery relative errors of the compared methods with respect to various sample ratios on GuangZhou and Portland dataset, respectively; (c) Visualization examples of imputation for GuangZhou dataset.
	}\label{traffic data}	
	\vspace{-4mm}
\end{figure}

\subsubsection{The Role of Similarity Scale}
On the GuangZhou dataset with $T = 60$, we evaluate the model’s adaptability to graph dynamics by varying the similarity scale $s$, and report the recovery errors in Figure~\ref{graph_ss} (c). The lowest errors are observed at $s = 5$ and $s = 10$, confirming the model’s ability to adjust to different levels of temporal variation via appropriate similarity scales. For the Portland dataset, due to its small $T$, we simply set $s = T$.

\subsubsection{Comparison with State-of-the-arts}
We compare the recovery performance of all methods on the two traffic datasets across varying sample ratios. The recovery relative errors with respect to the sample ratios are plotted in Figure~\ref{traffic data} (a)(b). Our method consistently outperforms the baselines, demonstrating superior recovery accuracy. Additionally, in Figure~\ref{traffic data} (c)  we showcase some visualization examples of the recovery results,  highlighting the high accuracy of our model in imputing plausible values from partially observed data.

\section{Conclusion}\label{section conclusion}
We propose a comprehensive framework for dynamic graph-regularized tensor completion, including a novel model, theoretical analysis, and efficient algorithm. At the modeling level, we introduce a rigorous mathematical formulation of dynamic graphs and derive a new tensor-oriented graph smoothness regularization, which effectively captures the global pairwise similarity structure within evolving graphs.
On the theoretical side, we show that the proposed regularization is equivalent to a weighted tensor nuclear norm, where the weights implicitly encode graph information. We further establish statistical consistency guarantees for our model, which is the first theoretical
guarantee  in the literature on graph-regularized tensor recovery.
Algorithmically, we design an efficient ADMM-based solver with convergence guarantees. Extensive experiments on both synthetic and real-world datasets demonstrate the superior recovery performance of our method, particularly under low sampling rates. Extending our framework to other graph regularized tensor recovery
problems, such as compressive sensing and robust PCA, presents an intriguing research direction.


%
%
%








\bibliographystyle{informs2014}
\bibliography{graphtensor_ref}

\end{document}